# A Method for Medical Data Analysis Using the LogNNet for Clinical Decision Support Systems and Edge Computing in Healthcare

Andrei Velichko

Institute of Physics and Technology, Petrozavodsk State University, 31 Lenina Str., 185910 Petrozavodsk,
Russia; velichko@petrsu.ru; Tel.: +7-9114005773

**Abstract:** Edge computing is a fast-growing and much needed technology in healthcare. The problem of implementing artificial intelligence on edge devices is the complexity and high resource intensity of the most known neural network data analysis methods and algorithms. The difficulty of implementing these methods on low-power microcontrollers with small memory size calls for the development of new effective algorithms for neural networks. This study presents a new method for analyzing medical data based on the LogNNet neural network, which uses chaotic mappings to transform input information. The method effectively solves classification problems and calculates risk factors for the presence of a disease in a patient according to a set of medical health indicators. The efficiency of LogNNet in assessing perinatal risk is illustrated on cardiotocogram data obtained from the UC Irvine machine learning repository. The classification accuracy reaches ~91% with the~3–10 kB of RAM used on the Arduino microcontroller. Using the LogNNet network trained on a publicly available database of the Israeli Ministry of Health, a service concept for COVID-19 express testing is provided. A classification accuracy of ~95% is achieved, and~0.6 kB of RAM is used. In all examples, the model is tested using standard classification quality metrics: precision, recall, and F1-measure. The LogNNet architecture allows the implementation of artificial intelligence on medical peripherals of the Internet of Things with low RAM resources and can be used in clinical decision support systems.

**Keywords:** clinical decision support systems; artificial intelligence; LogNNet; neural networks; edge computing; COVID-19; perinatal risk

## 1. Introduction

The Internet of Things (IoT) consists of intelligent devices that have limited resources and that are capable of collecting, recognizing, and processing data as well as exchanging processed data between network participants [1]. IoT is the backbone technology in various areas, including in smart healthcare [2,3], smart homes [4], smart urbanization [5]. The concept of smart healthcare is actively developing in different countries, and the global market for IoT medical devices is growing every year [6]. The constantly emerging new threats to public health, such as the new coronavirus disease 2019 (COVID-19) pandemic [7], crate continuous stimulus for the development of new technologies. As intelligent data processing requires the use of neural networks and intelligent algorithms, the concept of edge computing is actively developing [8–10]. In edge computing, part of the computing load is distributed between local devices with the connected sensors (Figure 1). Next, the processed information goes to the edge node for additional processing, and it then goes to the fog and cloud servers for global registration and processing. If the Internet connection with the fog and cloud servers fails, local intelligent processing helps to make a decision to solve the problem on the spot. The number of approaches to organize computing include the edge–cloud IoT model, the local edge–cloud IoT model [11], nanoEdge [12], and the software-defined network controller in the edge server [10]. Edge computing increases the efficiency of resources that are used by reducing the amount of data transferred between end systems and centralized cloud servers [9]. It enhances the communication and computing speed of IoT devices in the healthcare system.

Intelligent devices with limited resources [13], such as devices with a small amount of RAM (2–32 kB), a weak processor, and small dimensions, can receive various information about a patient's condition from different sensors. These sensors can be wearable sensors, which can measure heart rate, blood pressure, body temperature, and glucose levels, and external sensors are installed at control points. In addition, diagnostic information can be obtained as registered symptoms via mobile questionnaires, mobile devices, or public touch panels. In edge computing, the information that is received is pre-processed by artificial intelligence (AI) on the peripheral device, generating local healthcare services. Local data processing improves the confidentiality, security, and reliability of systems operating on the edge architecture [8].

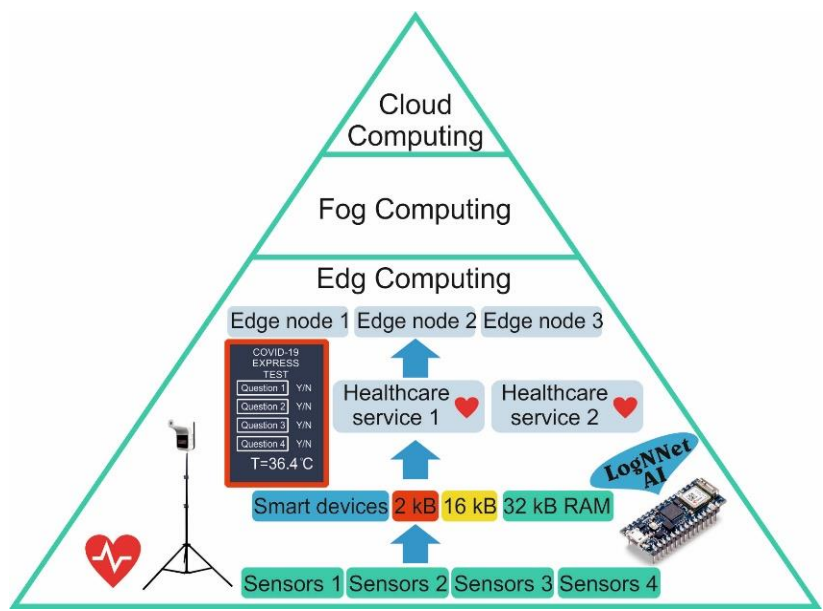

**Figure 1.** Architecture of IoT and edge computing in healthcare, powered by AI on LogNNet.

The key point is the presence of artificial intelligence on devices with limited resources. Therefore, the development of new neural architectures and algorithms capable of working on constrained devices is an important task.

Machine learning helps to create data processing rules for the purposes of clustering, classification, and regression, where unsupervised, supervised, and reinforced learning approaches are used [14]. Popular machine learning algorithms for analyzing medical data are multilayer perceptron (feedforward neural networks with multiple layers, linear classifier) [15], support vector machines [16], K-nearest neighbors [17], the random forest method [18], logistic regression [19], and decision trees [20,21]. Clinical decision support systems often use boosting methods, for example, AdaBoost [22] and the XGBoost classifier [23], when the algorithms are organized into assemblies to increase predictive accuracy. The disadvantage of this approach for edge computing is the lack of a clear understanding by users of how these algorithms are arranged, how to install them on operating systems with truncated libraries, how to reduce the memory and processing power they use, and how to implement them on microcontroller programming languages. Often, a task that can be solved on several neurons is approached using special libraries in the Python language and complex architectures with an excessive number of neurons and connections between them. This calls for the development of simple algorithms without complexity drawbacks, while having an efficiency comparable to that of known algorithms.

Ways to reduce the memory consumption of neural networks in edge computing [24] include pruning connections after learning [25,26], online learning with sparse networks [27], and quantization-aware training, which uses reduced bit precision per weight [28]. An additional method is proposed by the author of this paper for recalculating weights through chaotic mappings [29]. The study from [29] describes a classifier based on the neural network LogNNet using the example of handwritten digit recognition from the MNIST database. LogNNet has a simple structure and can operate on devices with low RAM (2–32 kB).

The operation principle of LogNNet is based on a matrix reservoir, which transforms the feature vector from the first multidimensional space into the second multidimensional space followed by the classification by a linear classifier. This transformation is the key for most machine learning methods, and the difference between the methods lies in the transformation algorithms, the number of algorithms, and its sequence. A distinctive attribute of LogNNet is the chaotic mixing of input features in various combinations in the reservoir, similar to the operation of reservoir neural networks [30] with the "kernel trick" effect [31]. This effect consists of increasing the efficiency of the linear classifier when the dimension of the feature space changes. Chaotic mapping allows an effective combination of features by selecting the optimal parameters in a vast area of chaotic states. Due to an effective combination of features in the reservoir and their subsequent classification, LogNNet has promising opportunities for application in clinical decision support systems.

The objective of this study is to create an effective method for analyzing medical data using the LogNNet neural network. The implementation possibilities of LogNNet on the peripheral devices of the medical IoT with low computing resources are demonstrated.

This paper has the following structure: Section 2 describes the basic LogNNet architecture followed by sections describing the method for using the neural network LogNNet for medical data analysis. The training, testing, and application steps for patient data analysis are detailed in flowcharts with text commentary. The final subsection provides an estimate of the RAM occupied by the neural network LogNNet for application in edge computing. Section 3 describes two examples using the methodology for assessing perinatal risk and presents the service for assessing the risk of COVID-19 disease by means of the calculation of basic classification metrics. Section 4 discusses the results and compares them with known developments. In the Conclusion, a general description of the study and its scientific significance is given.

## 2. Materials and Methods

One of the most important tasks that needs to be solved by artificial intelligence is the classification of an object represented as a vector of features. Almost any object or phenomenon can be associated with several features, which can be used to predict its behavior or relation to any class. For example, for classification problems using photographs, the feature vector is a vector of the pixels of the image color. For assessing the health of a patient, it is a vector of a set of symptoms and other medical indicators.

This section describes the general principle of operation of the LogNNet neural network and the method of using LogNNet in the classification of medical data.

There are two types of medical databases designed for training neural networks: Type 1 is divided into testing and training sets, and Type 2 is not divided into test and training sets. This paper provides examples of the use of both types of medical databases.

Clinical decision support system models were tested using the standard metrics for quality classification: precision, recall, and F1-measure, which is the harmonic mean between accuracy and completeness. For Type 1 databases, the neural network training was conducted on a training set, and testing was done on a test set. For Type 2 databases, the K-fold cross-validation technique was used [32] when all of the data was divided into $K$ parts (in this work, $K = 5$), and one of the parts was used as a test set, and the remaining $K$-1 parts were taken as the training set. Then, each of the $K$ parts of the database became a test set, and the average value of these metrics was calculated for all $K$ cases.

## 2.1. The Operation Principle of the Neural Network LogNNet

Figure 2a presents a general diagram of the LogNNet operation principle.

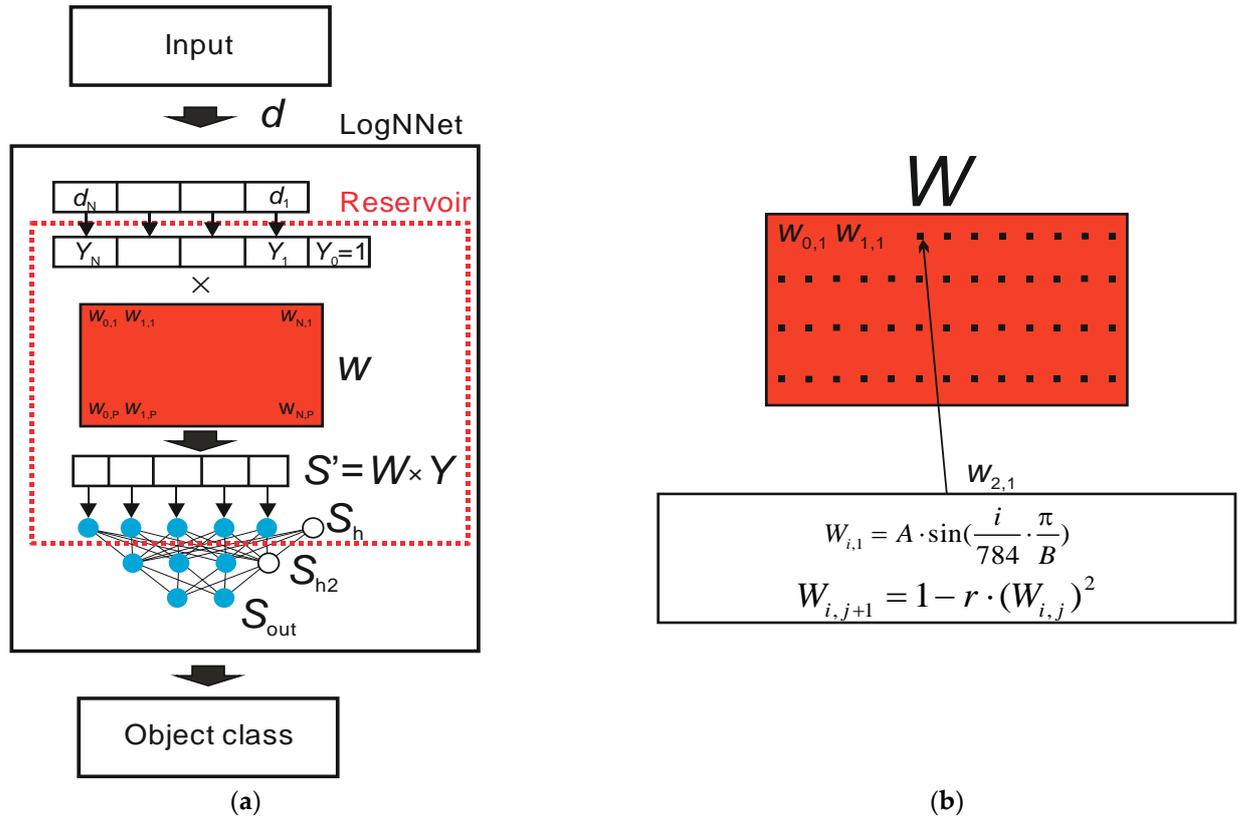

**Figure 2.** Block diagram of the neural network LogNNet (**a**). Block diagram of the filling process in the special matrix $W$ (**b**).

The input object in the form of a feature vector, denoted as $d$, enters the LogNNet classifier. The feature vector contains $N$ coordinates ($d_1, d_2… d_N$), where the number $N$ is determined by the user. For example, in study [29] on the handwritten digit recognition of MNIST database, the value $N$ equals 784, and it corresponds to the number of pixels in a 28 × 28 pixel image. At the output of the classifier, the object class of the input feature vector $d$ is determined. Let us denote the number of possible classes as $M$. In [29], the value $M$ corresponded to $M$ = 10, as the number of classes was determined by the value of the figure in the image from "0" to "9". Inside the classifier, there is a reservoir with a matrix, which is designated as $W$. First, vector d is transformed into a vector $Y$ of dimension $N + 1$ with an additional coordinate $Y_0 = 1$, and each component is normalized by dividing by the maximum value of this component in the training base. Second, matrix $W$ of dimension $(N + 1) \times P$ is multiplied by a vector $Y$. The result is a vector $S'$ with $P$ coordinates, which is normalized [29] and translated into a vector $S_h$ of dimension $P + 1$ with zero coordinate $S_h [0] = 1$. Zero coordinate acts as an offset element. Therefore, there is a transformation of the feature vector $d$ into the $(P + 1)$-dimensional space. Next, the vector $S_h$ is fed to a two-layer linear classifier with the number of neurons $H$ being fed into the hidden layer $S_{h2}$ and the number of outputs $M$ being fed into the output layer $S_{out}$.

A block diagram of the filling in the matrix $W$ is shown in Figure 2b. In [29], a filling method uses two equations. Based on the sine function, the first line of the $W$ matrix is filled according to the equation

$$W_{i,1} = A \cdot \sin(\frac{i}{784} \cdot \frac{\pi}{B}) \tag{1}$$

where parameter $i$ varies from 0 to $N$, parameter $A$ equals 0.3, and parameter $B$ equals 5.9.

Subsequent matrix elements are filled in according to the logistic mapping equation

$$W_{i,j+1} = 1 - r \cdot (W_{i,j})^2, \tag{2}$$

where $j$ ranges from 1 to $P$, and parameter $r$ ranges from 0 to 2.

The value of parameter *r* affects the classification accuracy of LogNNet [29], and the highest image classification accuracy is achieved when *r* corresponds to the region of chaotic behavior of the logistic mapping.

The training of the linear classifier LogNNet is performed by the back propagation method of the error [29].

*2.2. Method for Using the Neural Network LogNNet for Medical Data Analysis*

In the presented method, it is assumed that all of the objects from the training and test sets as well as the user data have the same dimension of the feature vector, that the coordinates of the vectors correspond to the same set and order of medical parameters, and that they do not contain missing values. A feature vector that does not meet these requirements must be removed from the database.

2.2.1. LogNNet Training

A block diagram of the LogNNet training process is shown in Figure 3. Training begins by retrieving a training set from a Type 1 or Type 2 database. Then, the training set is balanced. The balancing stage implies equalizing the number of objects for each class, supplementing the classes with copies of already existing objects, and sorting the training set in sequential order. The balancing process can be illustrated using an example. Let us suppose the training set consists of 10 objects. Each object is assigned a feature vector $d\_z$, where *z* is the object number $z = 1\ldots10$. All of the objects are divided into three classes. For example, we have five objects of Class 1 ($d\_1, d\_2, d\_4, d\_7, d\_10$), three objects of Class 2 ($d\_3, d\_8, d\_9$), and two objects of Class 3 ($d\_5, d\_6$). We find the maximum number of objects (*MAX*) in the classes, and in our example, *MAX* equals 5 for Class 1. We supplement the remaining groups with copies of the already existing objects (duplication) to equalize the number to *MAX*. Therefore, for Class 2, we acquire the group ($d\_3, d\_8, d\_9, d\_3, d\_8$), and for Class 3—($d\_5, d\_6, d\_5, d\_6, d\_5$). Then, we compose a balanced training data set, choosing one object from each group in turn. As a result, we achieve the following training set: ($d\_1, d\_3, d\_5, d\_2, d\_8, d\_6, d\_4, d\_9, d\_5, d\_7, d\_3, d\_6, d\_10, d\_8, d\_5$), which consists of 15 vectors and has the same number of objects in every class.

At the next stage, the values of the constant parameters of the model are set: value *P* determines the dimension of the vectors $S'$ and $S_h$, the number of layers in the linear classifier, the number of training epochs by the backpropagation method, and the number of neurons in the hidden layer of the classifier in the case of a two-layer classifier.

In addition, it is necessary to select a list of optimized parameters, which are the chaotic mapping parameters and other equations for the row-wise filling of the matrix *W*. In the basic version of LogNNet [29], these are parameters *A*, *B*, and *r* of Equations (1) and (2). In this study, additional chaotic mappings to fill the matrix *W* are used (see Table 1). In particular, the Henon map and its modification [33] were applied.

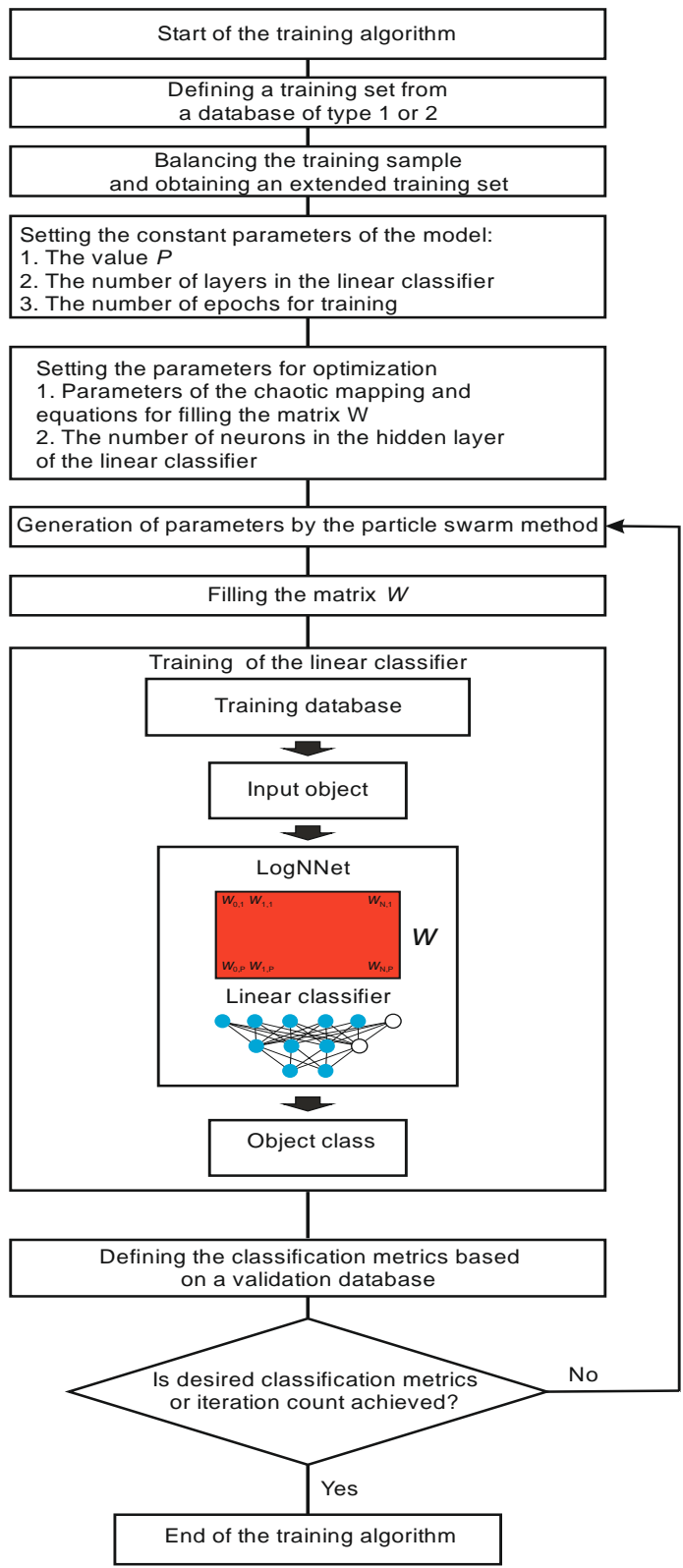

**Figure 3.** LogNNet training algorithm.

**Table 1.** Chaotic mappings, lists of optimized parameters with the limits of its variation, equations, and models.

| Chaotic Mapping Name | Model Name | List of Parameters to Optimize (Limits) | Equation |
|---|---|---|---|
| Sine and Logistic as in [29] | LogNNet | $A$ (−1 to 1) <br> $B$ (0.1 to 10) <br> $r$ (0.1 to 2) | Equations (1) and (2) |
| Logistic | LogNNet/Logistic | $x_0$ (−1 to 1) <br> $r$ (1 to 4) | $x_{n+1} = r \cdot x_n \cdot (1 - x_n)$ |
| Sine | LogNNet/Sine | $x_0$ (−1 to 1) <br> $r$ (0 to 2) | $x_{n+1} = r \cdot \sin(\pi \cdot x_n)$ |
| Gauss | LogNNet/Gauss | $x_0$ 0 (−1 to 1) <br> $r_1$ (−1 to 1) <br> $r_2$ (3 to 6) | $x_{n+1} = \exp(-r_2 \cdot x_n^2) + r_1$ |
| Two-sided | LogNNet/2sided | $x_0$ (0 to 10) <br> r (0 to 100) | $x_{n+1} = \dfrac{r \cdot x_n}{1 + x_n^3}$ |
| Plank | LogNNet/Plank | $x_0$ (0 to 5) <br> $r$ (0 to 7) | $x_{n+1} = \dfrac{r \cdot x_n^3}{1 + \exp(x_n)}$ |
| Henon | LogNNet/Henon1 | $x_0$ (0.01 to 1.5) <br> $r$ (0.01 to 10) <br> $r_1$ (0 to 1.5) <br> $r_2$ (0 to 1.5) | $\begin{cases} x_{n+1} = 1 - r_1 \cdot x_n^2 + y_n \\ y_{n+1} = r_2 \cdot x_n \end{cases}$ |
| Modified Henon | LogNNet/Henon2 | $x_0$ (0.01 to 1.5) <br> $y_0$ (0.01 to 10) <br> $r_1, r_2, r_3, r_4$ (0 to 1.5) | $\begin{cases} x_{n+1} = x_n + r_1 \cdot x_n^2 + r_2 \cdot y_n^2 - r_3 \cdot y_n \cdot x_n - r_4 \\ y_{n+1} = x_n \end{cases}$ |

The training of the LogNNet network begins with two nested iterations. The internal iteration trains the output LogNNet classifier by using the backpropagation method on the training set. The external iteration optimizes the model parameters and uses the particle swarm method. The variation limits of the optimized parameters (see Table 1), the constants of the optimization method, the weight fraction of inertia, and the local and global weight fractions are set. After setting the constants, the particle swarm algorithm generates the values of the model parameters, and the matrix $W$ is filled in. The filling is performed line by line, as shown in Figure 2b. The higher the entropy of the numerical series filling the special matrix, the better the classification accuracy [33,34]. Therefore, the procedure for optimizing the parameters of chaotic mapping plays an important role in the presented method for analyzing medical data using the LogNNet neural network.

After training the linear classifier, the classification metrics are determined based on the validation set, which, in general, is a training set.

Next, we check for exiting the optimization cycle of parameters. The exit occurs either when the desired values of the classification metrics are reached or when a given number of iterations in the particle swarm method is reached. If the condition is not satisfied, the next iteration occurs, and new model parameters are generated by the particle swarm method. If the condition is satisfied, the training algorithm ends.

As a result, we obtain the optimized parameters of the model (parameters of the chaotic mapping) at the output, which make it possible to obtain the highest classification accuracy possible on the validation set.

### 2.2.2. LogNNet Testing

The testing algorithm is shown in Figure 4. System testing begins with the operation of retrieving a test set from a Type 1 or Type 2 database. A prerequisite is that the test data should not participate in the training process described in the previous paragraph. The constant parameters of the model are set, corresponding to the same values as they do during training. Next, the parameters obtained after training the model are set: the parameters of the chaotic mapping, the equations for filling the matrix *W*, and the weight of the output classifier. A matrix *W* is filled in line by line, and the LogNNet network is tested on the test data. The classification metrics are defined, and the algorithm ends.

After the test data has been verified and the classification metrics meet the criteria for implementing the model in clinical practice, the model can be used to process patient data.

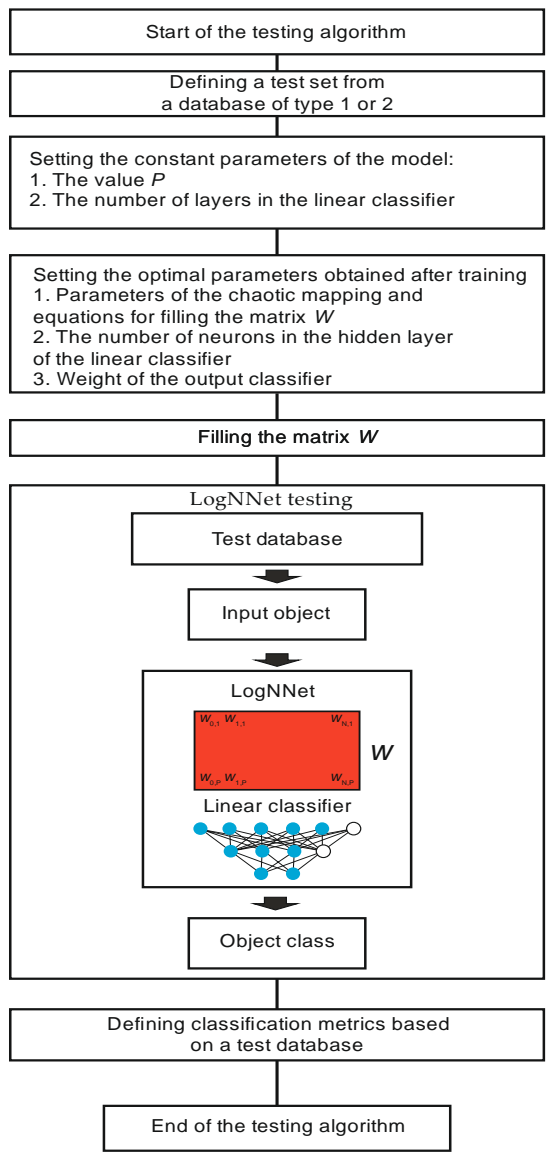

**Figure 4.** LogNNet testing algorithm.

2.2.3. Algorithm for Processing Patient's Medical Data Using LogNNet

The algorithm for processing a patient's data is shown in Figure 5. At the beginning, a feature vector of a patient's medical data is obtained. If the vector misses some data, this data is added. Next, a check is made for the compliance of the feature vector with the format that was used during training and testing. If the vector format corresponds to the model, then the analysis process begins, and if not, the vector is corrected.

Before the classification, constant values and optimal parameters are obtained from the training, the values of the classifier weights are set, and the matrix *W* is filled in. The analyzed vector is fed to the neural network LogNNet, and the object class is determined at the output. Based on the result, the risk factors are assessed for the presence of a disease in the patient, and the algorithm ends.

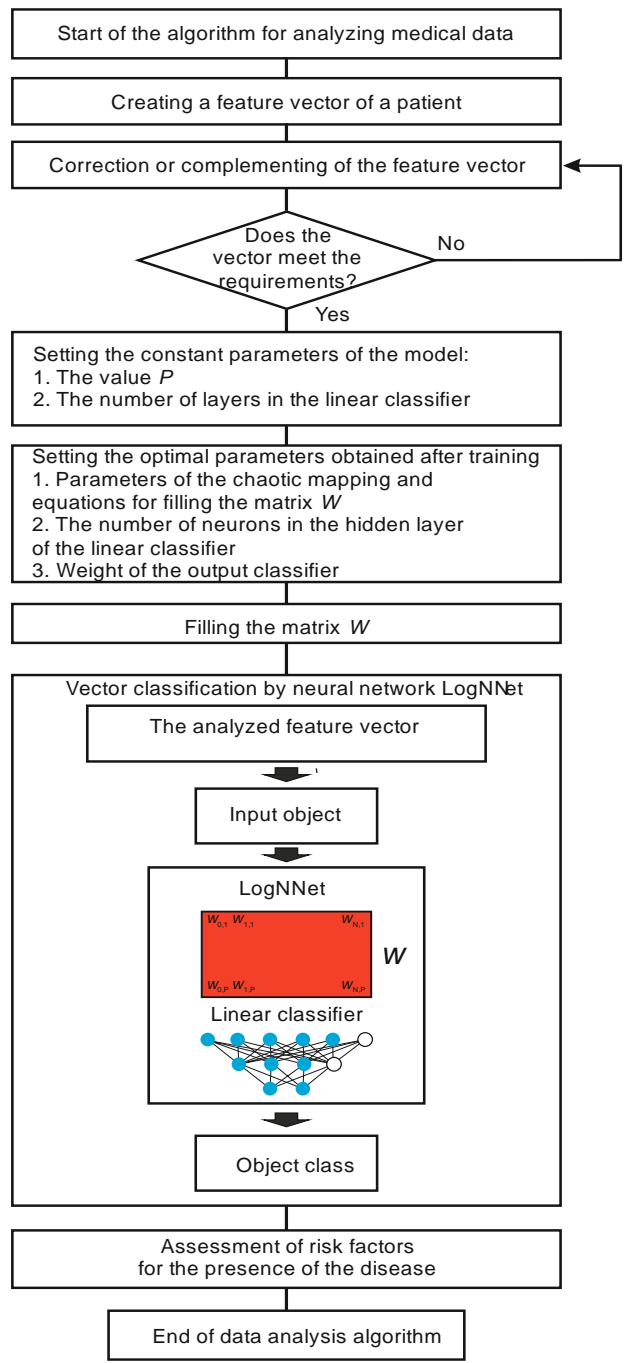

**Figure 5.** Algorithm for analyzing a patient's medical data using LogNNet.

*2.3. Estimation of the RAM Occupied by the Neural Network LogNNet for Application in Edge Computing*

To implement neural networks in edge computing, they should work on small micro-controllers with limited computing resources. The LogNNet network can effectively operate on boards of the Arduino family with a RAM size of up to 2 kb, and the successful results of recognizing handwritten digits from the MNIST database have been demonstrated [29,35].

Table 2 demonstrates the RAM consumption values on the Arduino microcontroller when implementing the LogNNet $N$:$P$:$H$:$M$ network with a two-layer classifier. The parameters $N$, $P$, $H$, and $M$ are described in Section 2.1.

The arrays used can be of "real" type (occupying 4 B) or "integer" type (occupying 2 B). Array $Y$ contains $(N + 1)$ "real" type elements. The "integer" array has the advantage of taking up less memory. The weights obtained from LogNNet training can be stored in an "integer" array [35].

**Table 2.** Estimation of the occupied RAM of the Arduino microcontroller when implementing the LogNNet $N$:$P$:$H$:$M$ network.

| Data Type | RAM Size |
|---|---|
| Array $Y$ | $(N + 1) \times 4$ B |
| Matrix $W$ | $(N + 1) \times P \times 4$ B |
| Weight matrix $S_h/S_{h2}$ | $(P + 1) \times H \times 2$ B |
| Weight matrix $S_{h2}/S_{out}$ | $(H + 1) \times M \times 2$ B |
| Array $S_h$ | $(P + 1) \times 4$ B |
| Array $S_{h2}$ | $(H + 1) \times 4$ B |
| Array $S_{out}$ | $M \times 4$ B |
| Auxiliary arrays | $P \times 3 \times 2$ B |
| Variables | 20 B |
| Serial port | 178 B |
| Memory for mathematical calculations | 200 B |

RAM saving is achieved by not storing the elements of the array $W$ in RAM, but by calculating each element during the operation of the network using the processor and the chaotic mapping equations from Table 1. The pseudo code for calculating $S' = W \cdot Y$ is given in Algorithm 1, where the elements of the matrix $W_{i,j}$ are replaced by the values $x_{n+1}$ of the chaotic mapping. Although this method reduces the speed of the neural network during object classification, it is not critical for practical implementation, as modern microprocessors are very fast. Algorithms without RAM saving (Algorithm 2) would allocate additional memory to store the array $W$ [24]. In Algorithm 2, line-by-line filling of the matrix $W$ with the chaotic mapping time series is performed before calculating $S'$.

**Algorithm 1.** LogNNet pseudo code without storing the elements of the array $W$ in RAM.

```
Transformation of input vector d into vector Y
// Calculation of the vector S'= W · Y
xn: = x0;
for j:=1 to P do
begin
  S'[j]:=0;
  for i:=0 to N do
  begin
    xnp1 = F(xn)
    S'[j]:=S'[j]+ xnp1*Y[i];
    xn: = xnp1;
  end;
end;
Calculation of the output layer of the reservoir S_h
Calculation of the output layer of the linear classifier S_out
Object class definition
```

**Algorithm 2.** LogNNet pseudo code with storing the elements of the array $W$ in RAM.

```
Transformation of input vector d into vector Y
// Line-by-line filling of the matrix W
xn: = x0;
for j: = 1 to P do
  for i: = 0 to N do
  begin
    xnp1 = F(xn)
    W[i,j]: = xnp1;
    xn: = xnp1;
  end;
end;
// Calculation of the vector S'= W · Y
```

```
                    for j:=1 to P do
                    begin
                      S'[j]:=0;
                      for i:=0 to N do S'[j]:=S'[j]+Y[i]*W[i,j];
                    end;
```
*Calculation of the output layer of the reservoir $S_h$*
*Calculation of the output layer of the linear classifier $S_{out}$*
*Object class definition*

The function F(xn) is one of the functions of the logistic mapping in Table 1, xn denotes $x_n$, xnp1 denotes $x_{n+1}$, and x0 is the initial value.

The LogNNet operational method allows the reduction of the size of the used RAM by the amount of memory allocated to the array *W* equal to (*N* + 1) × *P* × 4 B. For the configuration of the LogNNet 784:100:60:10 neural network described in [24], the amount of saved RAM can reach (784 + 1) × 100 × 4 ≈ 306 kB.

## 3. Results

This section presents the results of LogNNet application to two models: a perinatal risk assessment model and a risk assessment model for COVID-19 disease caused by the SARS-CoV-2 virus.

### *3.1. Perinatal Risk Assessment Model*

Complications during childbirth are one of the main causes of perinatal mortality [36–38]. A fetal cardiotocogram (CTG) can be used as a monitoring tool to identify high-risk women during childbirth [23]. In this example, the goal was to study the accuracy of the machine learning method based on the LogNNet neural network on CTG data when identifying women from high-risk groups. The CTG data of 2126 pregnant women were obtained from the UC Irvine Machine Learning Repository [39]. The database contains a set of features for each patient, presented in Table 3.

**Table 3.** List of features in the "cardiotocographic data" database.

| №   | Designation | Meaning |
| --- | --- | --- |
| 1.  | FileName | FileName of CTG examination |
| 2.  | Date | Date of the examination |
| 3.  | b | start instant |
| 4.  | e | end instant |
| 5.  | LBE | baseline value (medical expert) |
| 6.  | LB | baseline value (SisPorto) |
| 7.  | AC | accelerations (SisPorto) |
| 8.  | FM | foetal movement (SisPorto) |
| 9.  | UC | uterine contractions (SisPorto) |
| 10. | ASTV | percentage of time with abnormal short-term variability (SisPorto) |
| 11. | mSTV | mean value of short-term variability (SisPorto) |
| 12. | ALTV | percentage of time with abnormal long-term variability (SisPorto) |
| 13. | mLTV | mean value of long-term variability (SisPorto) |
| 14. | DL | light decelerations |
| 15. | DS | severe decelerations |
| 16. | DP | prolongued decelerations |
| 17. | DR | repetitive decelerations |
| 18. | Width | histogram width |
| 19. | Min | low freq. of the histogram |
| 20. | Max | high freq. of the histogram |
| 21. | Nmax | number of histogram peaks |
| 22. | Nzeros | number of histogram zeros |
| 23. | Mode | histogram mode |
| 24. | Mean | histogram mean |
| 25. | Median | histogram median |
| 26. | Variance | histogram variance |
| 27. | Tendency | histogram tendency: −1 = left asymmetric; 0 = symmetric; 1 = right asymmetric |

The output values for each patient are categorized into three risk categories:

(1) "N" (Normal);
(2) "S" (Suspicious);
(3) "P" (Pathology).

For the study, 25 features were selected: features 3–27. The first two field-files, name and date, did not participate in the network training process.

The presented database is Type 2 database. Therefore, during training and testing, the *K*-fold cross-validation method was used with *K* = 5.

Balancing was done separately for the training set, while the test set remained unchanged.

Two architectures were considered: LogNNet 25:100:40:3 and LogNNet 25:50:20:3, with different variants of chaotic mappings from Table 1. The values of the constant parameters of the model were set: the number of layers in the linear classifier is 2, the number of epochs for training is 50, the value *P* and the number of neurons in the hidden layer *H* of the linear classifier were determined by the architectures are *P* = 100, *H* = 40 (25:100:40:3) and *P* = 50, and *H* = 20 (25:50:20:3).

The model testing results are presented in Tables 4 and 5.

**Table 4.** Classification metrics of the perinatal risk assessment model for the LogNNet network architecture 25:100:40:3.

| Layer Architec-Ture 25:100:40:3 Model | Overall Accuracy, $A$, % | Precision | | | Recall | | | F1 | | |
|---|---|---|---|---|---|---|---|---|---|---|
| | | "N" | "S" | "P" | "N" | "S" | "P" | "N" | "S" | "P" |
| LogNNet | 90.935 | 0.997 | 0.688 | 0.873 | 0.889 | 0.969 | 0.989 | 0.939 | 0.795 | 0.919 |
| LogNNet/Logistic | 89.936 | 0.994 | 0.668 | 0.870 | 0.879 | 0.951 | 0.994 | 0.931 | 0.771 | 0.920 |
| LogNNet/Sine | 90.156 | 0.999 | 0.670 | 0.873 | 0.876 | 0.983 | 0.994 | 0.931 | 0.785 | 0.921 |
| LogNNet/Gauss | 90.313 | 0.994 | 0.663 | 0.909 | 0.883 | 0.958 | 0.994 | 0.933 | 0.771 | 0.943 |
| LogNNet/2sided | 82.745 | 0.981 | 0.562 | 0.797 | 0.800 | 0.889 | 0.965 | 0.875 | 0.667 | 0.851 |
| LogNNet/Plank | 91.177 | 0.993 | 0.700 | 0.876 | 0.896 | 0.943 | 0.994 | 0.940 | 0.794 | 0.922 |
| LogNNet/Henon1 | 91.190 | 0.999 | 0.674 | 0.899 | 0.891 | 0.980 | 0.994 | 0.941 | 0.791 | 0.936 |
| LogNNet/Henon2 | 90.683 | 0.999 | 0.674 | 0.891 | 0.883 | 0.983 | 0.994 | 0.936 | 0.790 | 0.928 |

**Table 5.** Classification metrics of the perinatal risk assessment model for the LogNNet network architecture 25:50:20:3.

| Layer Architec-Ture 25:50:20:3 Model | Overall Accuracy, $A$, % | Precision | | | Recall | | | F1 | | |
|---|---|---|---|---|---|---|---|---|---|---|
| | | "N" | "S" | "P" | "N" | "S" | "P" | "N" | "S" | "P" |
| LogNNet | 89.505 | 1.000 | 0.673 | 0.855 | 0.867 | 0.990 | 0.988 | 0.926 | 0.793 | 0.896 |
| LogNNet/Logistic | 89.048 | 0.990 | 0.663 | 0.826 | 0.875 | 0.916 | 0.983 | 0.927 | 0.754 | 0.888 |
| LogNNet/Sine | 90.080 | 0.996 | 0.667 | 0.861 | 0.880 | 0.965 | 0.982 | 0.932 | 0.778 | 0.909 |
| LogNNet/Gauss | 89.812 | 0.996 | 0.650 | 0.877 | 0.879 | 0.949 | 0.994 | 0.932 | 0.757 | 0.927 |
| LogNNet/2sided | 81.988 | 0.975 | 0.578 | 0.645 | 0.811 | 0.781 | 0.941 | 0.877 | 0.651 | 0.743 |
| LogNNet/Plank | 88.881 | 0.986 | 0.683 | 0.810 | 0.876 | 0.902 | 0.977 | 0.924 | 0.764 | 0.870 |
| LogNNet/Henon1 | 88.593 | 0.995 | 0.666 | 0.809 | 0.863 | 0.942 | 0.989 | 0.921 | 0.761 | 0.876 |
| LogNNet/Henon2 | 89.291 | 0.996 | 0.650 | 0.860 | 0.870 | 0.961 | 0.983 | 0.926 | 0.763 | 0.908 |

The models with a layer architecture of 25:100:40:3 demonstrated a high classification accuracy of about 90% (Table 4). The largest value, $A$ = 91.19%, corresponded to the LogNNet/Henon1 25:100:40:3 model, and the smallest value, $A$ = 82.745%, corresponded to the LogNNet/2sided 25:100:40:3 model. For the model with fewer neurons, with architecture 25:50:20:3 (Table 5) and Sine and Gaussian mappings, performed better, with $A$ = 90.080% and $A$ = 89.812%, respectively. For both architectures, the two-sided chaotic mapping (LogNNet/2sided model) showed the worst result.

*3.2. Model for Assessing the Risk of COVID-19 Disease Caused by the SARS-CoV-2 Virus*

The new coronavirus disease 2019 (COVID-19) pandemic caused by SARS-CoV-2 continues to pose a serious threat to public health [40]. The ability to make clinical decisions quickly and the ability to use health resources efficiently is essential in the fight against the pandemic. One of the modern testing methods for coronavirus infection is PCR analysis using polymerase chain reaction [41]. Analysis availability has long been difficult in developing countries, which has contributed to increased infection rates. Therefore, the development of effective screening methods that can quickly diagnose COVID-19 and that can help reduce the burden on health systems is an important direction in the development of medical diagnostic methods. Even with available PCR analysis, not many people go to diagnostic laboratories due to the general workload and poor awareness of the signs of the disease. It is important to develop predictive models for the COVID-19 test results. The models are designed to help medical personnel in patient triage, especially in conditions with limited health resources, and to promote the development of mobile services for self-diagnosis at home.

The Israeli Ministry of Health has published data on individuals who have been tested for SARS-CoV-2 using a nasopharyngeal swab by means of PCR [42]. These data are actively used by scientists to create forecasting models [43]. In addition to the date and result of the PCR test, various information is available in the initial database, including clinical symptoms, gender, information as to whether the person is over 60 years of age, and whether the person has had contact with an infected person. The list of fields is given in Table 6. Information can be presented in the form of answers (Yes or No) to the questions posed or in binary form (0 or 1). Clinical symptoms can be obtained during the initial examination of the patient. The procedure does not require significant medical center resources. The patient can be interviewed at home, or self-examination is used.

**Table 6.** Patient data in the database.

| № | Fields | Answer Options | Binary form for LogNNet Model |
|---|---|---|---|
| 1 | Sex | Male/Female | 1/0 |
| 2 | Age is ≥ 60 | Yes/No | 1/0 |
| 3 | Cough | Yes/No | 1/0 |
| 4 | Fever, high temperature | Yes/No | 1/0 |
| 5 | A sore throat | Yes/No | 1/0 |
| 6 | Dyspnea | Yes/No | 1/0 |
| 7 | Headache | Yes/No | 1/0 |
| 8 | Contact with a person who has confirmed COVID-19 | Yes/No | 1/0 |
| 9 | PCR Test result for COVID-19 | Negative/Positive | 0/1 |
| 10 | Date | | |

Based on this data, a LogNNet model was developed to predict COVID-19 test results using the eight binary characteristics presented in Table 6 under numbers 1–8.

The database is classified as a Type 1 database, where the training and test sets are defined, similar to [43]. The training set consisted of 46,872 records of individuals who had been tested, where 3874 cases were COVID-19 (positive) and 42,998 cases were not confirmed (negative) for the period from 22 March 2020 to

31 March 2020. The test set contained data for the next week, from 1–7 April 2020, and consisted of 43,916 who had been individuals tested, where 3370 cases were confirmed to have COVID-19.

In accordance with the algorithm (Figure 3), the training set was balanced, and the sample size increased to 85,996. The number of confirmed and unconfirmed COVID-19 diagnoses leveled off at 42,998. Next, the constant parameter values of the model were set: the number of layers in the linear classifier was equal to 2, the number of training epochs was 50, and the value $P$ and the number of neurons in the hidden layer $H$ of the linear classifier were determined by two architectures 8:16:10:2 ($P = 16$, $H = 10$) and 8:6:4:2 ($P = 6$, $H = 4$). All of the chaotic mappings presented in Table 1 were tested. Further, the parameters were optimized by the particle swarm method. After finding the optimal values, testing was performed as in algorithm in Figure 4. The testing results for the 8:16:10:2 and 8:6:4:2 architecture models are presented in Tables 7 and 8.

**Table 7.** Model classification metrics for assessing the risk of COVID-19 disease under various chaotic mappings for LogNNet architecture 8:16:10:2.

| Layer Architec- Ture 8:16:10:2 Model | Overall Accuracy, $A$, % | Precision | | Recall | | F1 | |
|---|---|---|---|---|---|---|---|
| | | Negative | Positive | Negative | Positive | Negative | Positive |
| LogNNet | 95.12 | 0.981 | 0.653 | 0.966 | 0.774 | 0.973 | 0.709 |
| LogNNet/Logistic | 94.93 | 0.981 | 0.641 | 0.964 | 0.771 | 0.972 | 0.700 |
| LogNNet/Sine | 95.12 | 0.981 | 0.653 | 0.966 | 0.774 | 0.973 | 0.709 |
| LogNNet/Gauss | 94.98 | 0.980 | 0.646 | 0.965 | 0.766 | 0.973 | 0.701 |
| LogNNet/2sided | 95.23 | 0.980 | 0.666 | 0.968 | 0.758 | 0.974 | 0.709 |
| LogNNet/Plank | 94.93 | 0.981 | 0.641 | 0.964 | 0.771 | 0.972 | 0.700 |
| LogNNet/Henon1 | 95.26 | 0.980 | 0.667 | 0.968 | 0.763 | 0.974 | 0.712 |
| LogNNet/Henon2 | 95.12 | 0.981 | 0.653 | 0.966 | 0.774 | 0.973 | 0.709 |

**Table 8.** Model classification metrics for assessing the risk of COVID-19 disease under various chaotic mappings for LogNNet architecture 8:6:4:2.

| Layer Architec- Ture 8:6:4:2 Model | Overall Accuracy, $A$, % | Precision | | Recall | | F1 | |
|---|---|---|---|---|---|---|---|
| | | Negative | Positive | Negative | Positive | Negative | Positive |
| LogNNet | 95.15 | 0.981 | 0.657 | 0.967 | 0.769 | 0.974 | 0.709 |
| LogNNet/Logistic | 95.12 | 0.981 | 0.653 | 0.966 | 0.774 | 0.973 | 0.709 |
| LogNNet/Sine | 95.46 | 0.975 | 0.707 | 0.976 | 0.698 | 0.975 | 0.702 |
| LogNNet/Gauss | 95.16 | 0.979 | 0.664 | 0.968 | 0.748 | 0.974 | 0.703 |
| LogNNet/2sided | 95.20 | 0.979 | 0.665 | 0.968 | 0.754 | 0.974 | 0.707 |
| LogNNet/Plank | 94.93 | 0.981 | 0.641 | 0.964 | 0.771 | 0.972 | 0.700 |
| LogNNet/Henon1 | 95.12 | 0.981 | 0.653 | 0.966 | 0.774 | 0.973 | 0.709 |
| LogNNet/Henon2 | 95.22 | 0.974 | 0.689 | 0.974 | 0.685 | 0.974 | 0.687 |

Tables 7 and 8 show the architectures with different numbers of neurons in reservoirs −16 and 8 and with a different number of neurons in the hidden layer of the output classifiers −10 and 4. The number of output neurons was two, which corresponds to the two classes of the disease presence (Negative and Positive). The network with the lower number of layers, 8:6:4:2, performs slightly better than the network with the 8:16:10:2 architecture. The best result corresponds to the LogNNet/Sine 8:6:4:2 model ($A = 95.46\%$) with chaotic sine mapping. The fact that the model with fewer neurons performed better is not a general rule, but rather an exception to the rule, since it is more difficult to retrain a system with fewer neurons. In addition, the input data are presented in binary form, and the amount of data is much less than in the first example. All of these factors can lead to the result that a small number of neurons can optimally solve this problem. Therefore, for each individual practical task, it is necessary to test several LogNNet architectures with a different number of neurons in the reservoir and in the hidden layer of the classifier and to choose the best architecture or include these parameters to be optimized along with the parameters of the chaotic mapping.

## 3.3. Estimation of RAM Occupied for the Arduino Microcontroller

For the different LogNNet architectures discussed above, the occupied RAM was estimated for implementation on Arduino microcontrollers. Table 9 demonstrates the RAM values for Algorithm 1 with RAM saving and for Algorithm 2 without RAM saving.

**Table 9.** Estimation of occupied ram for the Arduino microcontroller when implementing the LogNNet network.

| LogNNet Architecture | RAM Algorithm 1 | RAM Algorithm 2 | RAM Saving |
|---|---|---|---|
| LogNNet 25:100:40:3 | 10,008 B | 20,408 B | 10,400 B |
| LogNNet 25:50:20:3 | 3268 B | 8468 B | 5200 B |
| LogNNet 8:16:10:2 | 1034 B | 1610 B | 576 B |
| LogNNet 8:6:4:2 | 602 B | 818 B | 216 B |

The LogNNet 25:50:20:3 model for assessing perinatal risk uses about 8 kB of RAM, and a more detailed RAM distribution is shown in Figure 6a. A significant part of the memory, ~5 kB, is occupied by the matrix *W*. When using Algorithm 1, this memory can be freed, and the algorithm will use only about 3 kB. The second biggest memory consumer (2 kB) is the array of weights between the Sh/Sh2 layers, as it contains the weights obtained after training the neural network. This neural network can be implemented on microcontrollers with 16–32 kB of memory, for example, Arduino Nano.

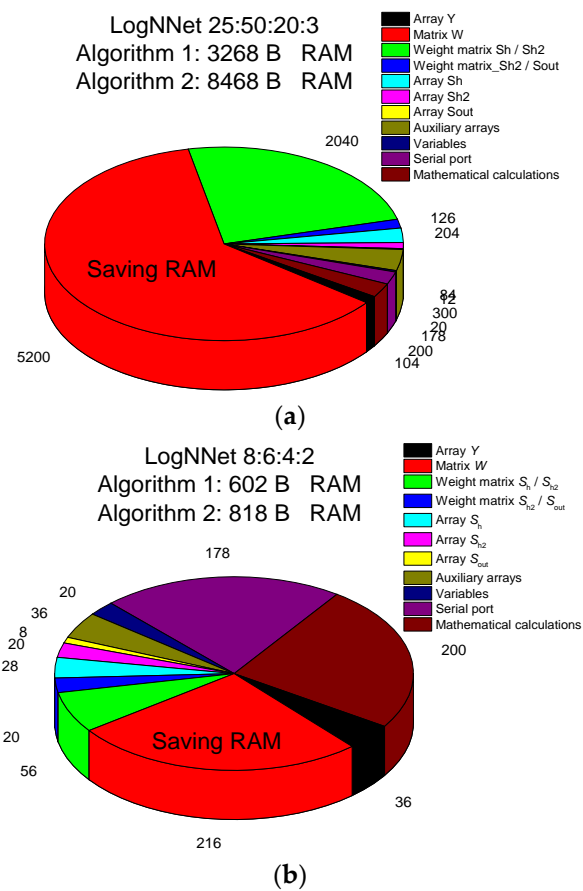

**Figure 6.** RAM memory allocation in the LogNNet 25:50:20:3 (**a**) and LogNNet 8:6:4:2 (**b**) configurations indicating the total memory used for Algorithms 1 and 2.

The LogNNet 8:6:4:2 model for estimating the risk of COVID-19 disease takes about 818 B of RAM (Figure 6b). As the matrix *W* occupies ~216 B, if Algorithm 1 is used, this memory can be freed, and the algorithm will use ~600 B. Therefore, the model can be placed on microcontrollers with a RAM size of 1–2 kB, for example, Arduino Uno. The RAM savings are not as significant as they are in the LogNNet 25:50:20:3 configuration because of the small number of neurons in the model.

## 4. Discussion

The algorithm for processing patient data is shown in Figure 5. However, not all patients have the required number of health indicators. In the absence of a particular indicator, the classifiers of the systems can malfunction, and it can lead to critical classification errors and errors in assessing the risk of a disease. The simplest method, when the missing data are replaced by the average value for a given indicator from the training database, leads to classification errors. The patent proposed in [44] proposes a method, where, in the absence of certain coordinates in the input feature vector, a classifier is used that is specifically trained for the features that are present. Therefore, it is necessary to either prepare several classifiers in advance and apply them depending on the presence of certain health indicators or to re-train the classifier on the fly, leaving only those features of health that are present for the patient in the training database. Testing a similar method on LogNNEt can be a topic for future research.

An increase in the space dimension of the input features occurs in the LogNNet reservoir. For example, in the LogNNet 25:100:40:3 configuration, the (25 + 1) -dimensional vector $Y$ is transformed into the (100 + 1) -dimensional vector $S_h$, which is then classified by the linear classifier. The more complex the chaos, the more diverse the $Y$ transformation in the reservoir, and the better the linear classifier classifies $S_h$ in the 101 dimensional space. In the LogNNet 8:6:4:2 configuration, the space dimension of the of input features in the reservoir is reduced, and the (8 + 1) -dimensional vector $Y$ is transformed into the (6 + 1) -dimensional vector $S_h$. This approach provides good results for tasks such as assessing the risk of COVID-19 (see Section 3.2) or MNIST image recognition [29]; as the main features are distinguished in the reservoir, the minor features are erased, and the output classifier is trained more efficiently.

Previous studies revealed [29,34] that the higher the Lyapunov exponent or the entropy of a chaotic mapping, the higher the accuracy of the LogNNet classification. Despite this finding, the present results call for the optimizing of the chaotic mapping parameters in the LogNNet reservoir to increase the efficiency of transforming the space of the input features. For each specific network configuration and classification problem, it is necessary to perform the separate optimization and selection of the mapping type to find the best solution. Under the same initial conditions, the chaotic mapping generates repeating time series. It is more advantageous than using a random number generator, as the chaotic dynamics in the optimization process can be varied using the control parameters of the mapping. The study of the role of chaotic mappings and chaos parameters in the transformation of input features in reservoir neural networks can be a topic for future research.

The presented model LogNNet/Henon1 25:100:40:3 (Table 4) for assessing perinatal risk has significantly better performance in terms of precision, recall, and F1 metrics for test data than the MLP multilayer perceptron model in [23]. The achieved accuracy A = 91.19% exceeds the results of a number of algorithms in similar studies: MLP ($A \approx 83\%$ [23]), logistic regression ($A \approx 82\%$ [23]), naive Bayes ($A \approx 81.56\%$ [45], 77% [23]), radial basis function networks ($A \approx 85.98\%$) [45], Bayesian networks ($A \approx 86.78\%$) [45], support vector machines ($A \approx 88.75\%$ [45], 90.64% [46], 90.65% [47]). Comparison with the XGBoost [23] model shows that the overall accuracy is at the same level of~91–92%, while LogNNet/Henon1 even surpasses the XGBoost classifier for individual metrics within the same class. Boosting algorithms (AdaBoost, XGBBoost) and LightGBM [48] allow higher accuracy values $A \approx 91\%-95\%$ to be obtained [45,49]; however, the advantage of the presented method is the possibility of implementing LogNNet on low-power IoT peripherals. The occupied RAM memory for LogNNet 25:100:40:3 is about 10 kB (Table 9), and it allows the implementation of the model on the Arduino Nano microcontroller. Lay health workers in low- and middle-income countries can use this model to triage pregnant women in remote areas for early referral and follow-up treatment.

The results of the model analysis for assessing the risk of COVID-19 disease showed that the use of almost all chaotic mappings led to a good prediction accuracy of about 95%, while the precision, recall, and F1 indicators were higher in the Negative class. The test results were at the level of [43], where the gradient-boosting predictor trained with the LightGBM was used. The best performance was demonstrated by the LogNNet/Sine 8:6: 4:2 ($A$ = 95.46%) model, with a chaotic sine mapping and 600 B of RAM occupied. This model can be placed on Arduino Uno microcontrollers with a RAM size of 2 kB. A service concept for the advanced medical diagnosis of COVID-19 can be proposed (see Figure 7). The technical part contains an Arduino Uno or Arduino nano board, with a connected temperature sensor and a touch panel. The patient is asked questions 1–8 from Table 6, which are displayed on a touch panel, and the patient's temperature is measured. At the output, the system evaluates the presence of the risk of COVID-19 disease. Currently, for example, in Thailand, the service of installing temperature sensors in public places is widespread. It produces a preliminary screening method for the temperatures of visitors. Such temperature modules can be equipped with artificial intelligence based on LogNNet and can offer better express testing for COVID-19.

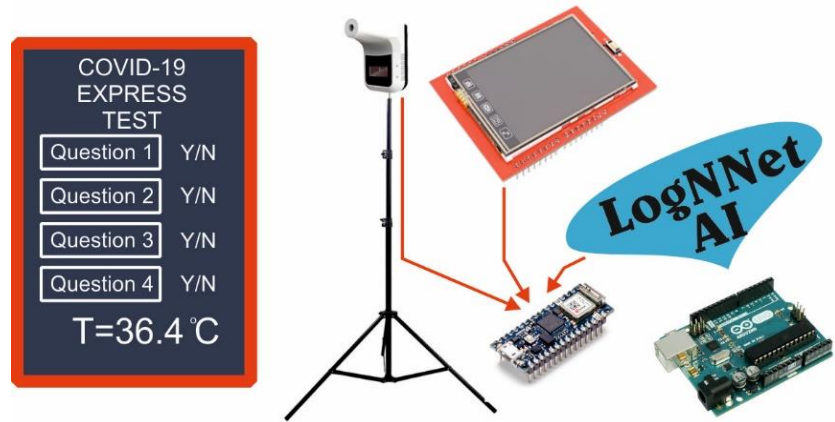

**Figure 7.** Service concept for preliminary medical diagnosis of COVID-19.

The results of the present study are in line with the concept of nanoEdge [12], where low-power devices at the nodes can provide collective services and process information together, communicating with each other using low-power wireless radios such as BLE, ZigBee, LoRa, or other similar technologies. In the field of health care, special attention is paid to low-energy and communication devices [50], and this highlights the need to develop neural network algorithms for constrained devices.

Endeavors to create resource-efficient algorithms and neural networks are actively being undertaken in the scientific community. For example, the algorithms Bonsai [51], ProtoNN [52], CNN [53], FastGRNN [54], Spectral-RNN [55], and NeuralNet Pruning [56] can be run for MNIST recognition tasks on devices with RAM in the 2–16 kB range. However, these algorithms have increased coding complexity, and there is no information on their effectiveness in analyzing medical data.

The results of this study open opportunities for the use of neural LogNNet models for mobile diagnostics in clinical decision support systems and in patient self-diagnostics. The described technique is universal and can be tested on a wide variety of medical databases. Further theoretical developments and the practical implementation of mobile health services and edge computing open wide perspectives for future research.

## 5. Conclusions

This paper presents a new algorithm for the implementation of a neural network based on the LogNNet architecture, which has shown its effectiveness in solving problems related to the classification of medical data. The role of chaos is highlighted, and the optimization of chaos properties in the process of transforming the space of input features in the reservoir affects the efficiency of the LogNNet classification. The algorithm paves the way for the development of edge computing in healthcare. The different types of algorithms and neural architectures are dedicated to effectively solve a certain class of problems. The presented algorithm is an advanced classification algorithm for clinical decision support systems, operating on low power microcontrollers with small memory size.

A method for medical data analysis using the LogNNet neural network is presented to calculate risk factors for the presence of a disease in a patient based on a set of medical health indicators. The algorithm illustrates the diagnosis of COVID-19 after training the LogNNet network using a publicly available database from the Israeli Ministry of Health. This database publishes data on the patients who have been tested for SARS-CoV-2 using a nasopharyngeal swab analysis by means of the PCR method. In addition, the LogNNet network assesses perinatal risk based on cardiogram data of 2126 pregnant women obtained from the machine learning repository of the University of California, Irvine. In all examples, the model was tested by evaluating standard classification quality metrics: precision, recall and F1-measure.

The results of this study can help to implement artificial intelligence on medical peripheral devices of the Internet of Things with low RAM resources, including clinical decision support systems, remote Internet medicine, and telemedicine.

## 6. Patents

An application for invention No. 2021117058 "Method for analyzing medical data using the neural network LogNNet" has been filed.

**Funding:** This research received no external funding.

**Institutional Review Board Statement:** Ethical review and approval were waived for this study. The Tel-Aviv University review board (IRB) determined that the Israeli Ministry of Health public dataset used in this study does not require IRB approval for analysis. Therefore, the IRB determined that this study is exempted from an approval.

**Informed Consent Statement:** Not applicable.

**Data Availability Statement:** All the data used in this study were retrieved from the Israeli Ministry of Health [42] and the UC Irvine machine learning repository [39] websites.

**Acknowledgments:** The author expresses his gratitude to Andrei Rikkiev for the valuable comments in the course of the article's translation and revision. Special thanks to the editors of the journal and to the anonymous reviewers for their constructive criticism and improvement suggestions.

**Conflicts of Interest:** The author declares no conflicts of interest.